\documentclass{article}
\usepackage{arxiv}

\usepackage[utf8]{inputenc} 
\usepackage[T1]{fontenc}    
\usepackage{amsmath}      
\usepackage[hyphens]{url}        
\usepackage{graphicx}
\RequirePackage{filecontents}    
\usepackage{booktabs}       
\usepackage{amsfonts} 
\usepackage{booktabs}
\usepackage{nicefrac}       
\usepackage{microtype}      
\usepackage{lipsum}
\usepackage{fancyhdr}       
\usepackage[table]{xcolor}
\usepackage{authblk}
\usepackage{doi}
\usepackage[numbers]{natbib}
\usepackage{hyperref}
\usepackage{cleveref}

\title{Strategies for Optimizing End-to-End Artificial Intelligence Pipelines on Intel\textsuperscript{\textregistered} Xeon\textsuperscript{\textregistered} Processors}

\renewcommand * {\Authfont}{\bfseries}
\author{\thanks{meena.arunachalam@intel.com}
	{Meena Arunachalam, Vrushabh Sanghavi, Yi A Yao, Yi A Zhou, Lifeng A Wang, 
		Zongru Wen, Niroop Ammbashankar, Ning W Wang, Fahim Mohammad}}
\affil {Artificial Intelligence and Analytics Group \\ 
	Intel Corporation} 

\renewcommand{\shorttitle}{Strategies for Optimizing E2E AI Pipelines.}

\hypersetup{
	pdftitle={Strategies for Optimizing End-to-End Artificial Intelligence Pipelines on Intel\textsuperscript{\textregistered} Xeon\textsuperscript{\textregistered} Processors},
	pdfsubject={cs.AI, cs.CV, cs.CL },
	pdfauthor={Meena Arunachalam, Vrushabh Sanghavi, Yi A Yao, Yi A Zhou, Lifeng A Wang, 
			Zongru Wen, Niroop Ammbashankar, Ning W Wang, Fahim Mohammad},
		pdfkeywords={Artificial Intelligence, Deep Learning Optimization, Optimization strategies, End-to-End AI applications, E2E performance optimization,  Intel\textsuperscript{\textregistered} Xeon\textsuperscript{\textregistered}}
	}

\begin{document}
	
	\maketitle
	
	\begin{abstract}
		End-to-end (E2E) artificial intelligence (AI) pipelines are composed of several stages including data preprocessing, data ingestion, defining and training the model, hyperparameter optimization, deployment, inference, postprocessing, followed by downstream analyses. To obtain efficient E2E workflow, it is required to optimize almost all the stages of pipeline. Intel\textsuperscript{\textregistered} Xeon\textsuperscript{\textregistered} processors come with large memory capacities, bundled with AI acceleration (e.g., Intel Deep Learning Boost), well suited to run multiple instances of training and inference pipelines in parallel and has low total cost of ownership (TCO). To showcase the performance on Xeon processors, we applied comprehensive optimization strategies coupled with software and hardware acceleration on variety of E2E pipelines in the areas of  Computer Vision, NLP, Recommendation systems, etc. We were able to achieve a performance improvement, ranging from 1.8x to 81.7x across different E2E pipelines. In this paper, we will be highlighting the optimization strategies adopted by us to achieve this performance on Intel\textsuperscript{\textregistered} Xeon\textsuperscript{\textregistered} processors with a set of eight different E2E pipelines.
	\end{abstract}


\section{Introduction}

	End-to-end AI pipelines are made up of one or more machine learning (ML)/deep learning (DL) models that solve a problem on specific dataset and modality, accompanied by multiple preprocessing and postprocessing stages. We apply comprehensive optimization strategies on a variety of modern AI pipelines including natural language processing (NLP), recommendation systems, video analytics, anomaly detection, and face recognition, along with DL/ML model training and inference, optimized data ingestion, feature engineering, media codecs, tokenization, etc. to achieve higher E2E performance. The results across all our candidate pipelines, mostly inference-based, show that for optimal E2E throughput performance, all phases must be optimized. Large memory capacities, AI acceleration (e.g., Intel Deep Learning Boost), and the ability to run general-purpose code make Xeon processors well-suited for these pipelines.
	
	Our optimizations fall broadly into application, framework, and library software; model hyperparameters; model optimization; system-level tuning; and workload partitioning. Tools such as Intel Neural Compressor (INC) offer quantization, distillation, pruning, and other techniques that benefit from DL Boost and other AI acceleration built into Intel\textsuperscript{\textregistered} Xeon\textsuperscript{\textregistered} processors. As a result, we see 1.8x to 81.7x improvement across different E2E pipelines. In addition, we can host multiple parallel streams or instances of these pipelines with the high number of cores and memory capacity available on Xeon compared to some memory-limited accelerators that can only host one or a very limited number of parallel streams. In many cases, workload consolidation to CPUs is possible, which also has TCO and power advantages.

\section{E2E AI Applications}
	\label{sec:e2eaiapp}
	We showcase many E2E AI use-cases and workloads, each comprised of unique pre/post processing steps and implemented using a variety of different ML/DL approaches on video, image, tabular, text, and other data types. (See Table \ref{tab:Table1}).
			 
			\begin{table}
				\centering
				\includegraphics[width=\textwidth]{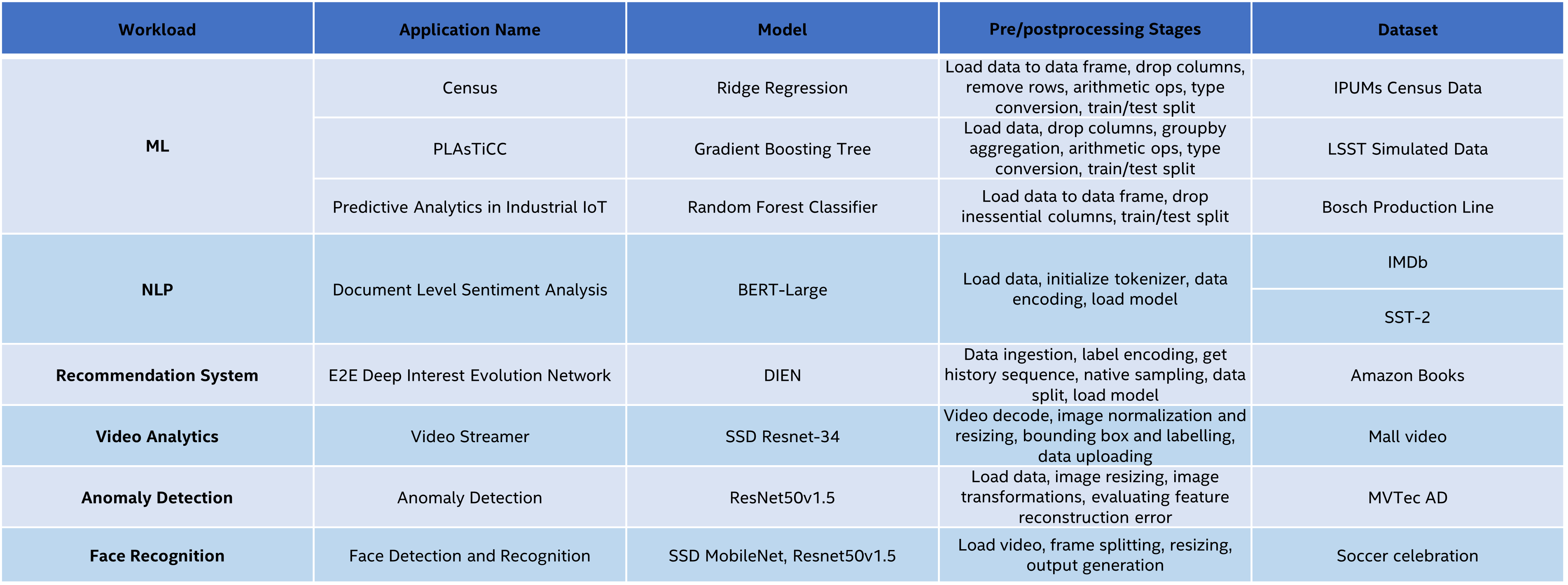}
				\caption{E2E AI applications.}
				\label{tab:Table1}
			\end{table}

			\begin{figure}
				\centering
				\includegraphics[width=\textwidth]{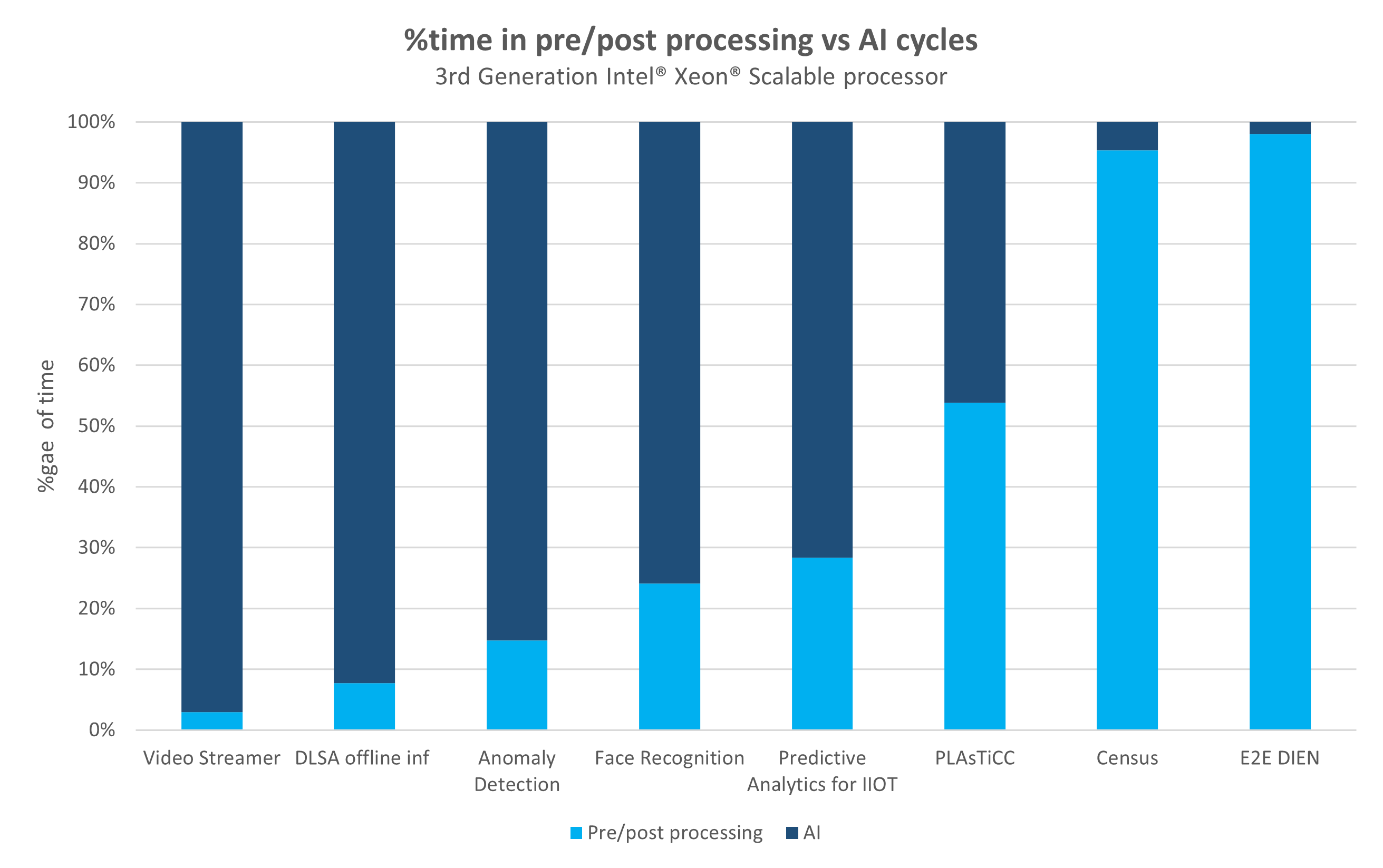}
				\caption{Percent time in pre/postprocessing vs AI.}
				\label{fig:fig1}
			\end{figure}

	E2E AI applications typically involve two broad categories of operations: 1) pre/post processing and 2) AI. We calculated the percentage of time spent on these two categories while carrying out E2E analyses. In Figure ~\ref{fig:fig1}, we see the breakdown range from 4\% to 98\% pre/postprocessing to 2\% to 96\% AI as a fraction of the total E2E run-time. Looking at these breakdowns, it becomes more compelling to optimize every stages in the pipeline. This section briefly presents a list workloads we developed and optimized. We also demonstrate how the components in these workload are together to get the best performance.

		\subsection{Census}
\begin{figure}
					\centering
					\includegraphics[width=\textwidth]{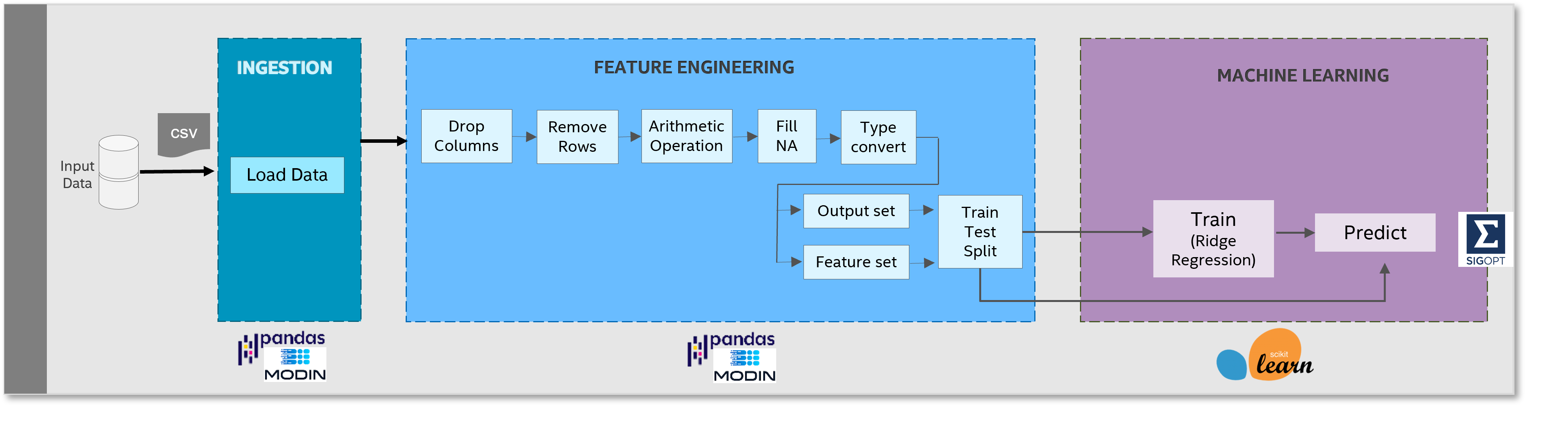}
					\caption{Census application pipeline.}
					\label{fig:fig2}
				\end{figure}

		The Census workload trains a ridge-regression  model using the US census data from the years 1970 to 2010 and predicts the correlation between personal education level and income (Figure \ref{fig:fig2}). Prior to ML, it ingests the data, performs data frame operations to prepare the input for model training and creates a feature set and its subsequent output set \cite{censuswl}. This analyses is based on IPUMS data \cite{ipums_data}. A more detailed analyses can be found here \cite{census_meena}.

		\subsection{PLAsTiCC}
\begin{figure}
					\centering
					\includegraphics[width=\textwidth]{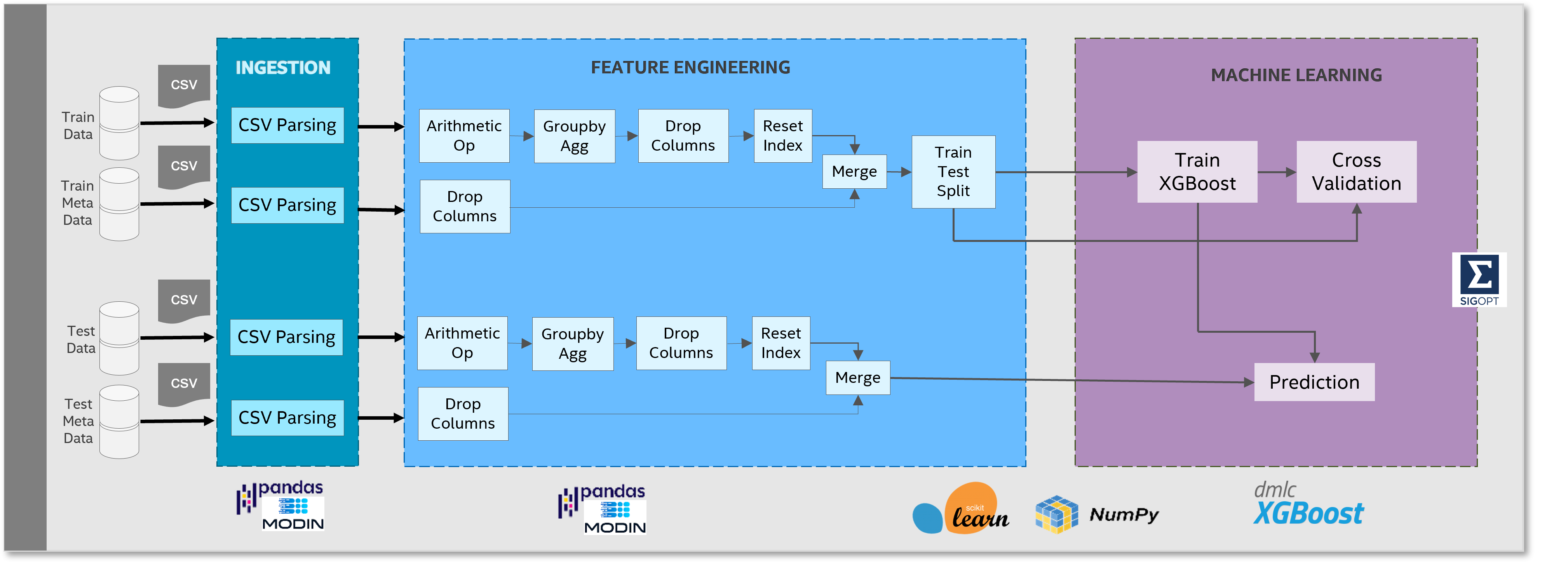}
					\caption{PLAsTiCC application pipeline.}
					\label{fig:fig3}
				\end{figure}

		The Photometric LSST Astronomical Time-Series Classification Challenge (PLAsTiCC) \cite{plasticcpaper} is an open data challenge that uses simulated astronomical time-series data to classify objects in the night sky that vary in brightness (Figure \ref{fig:fig3}). This pipeline loads the data; manipulates, transforms, and processes the data frames; and uses the histogram tree method from the XGBoost \cite{sklearn-bench, Chen-2016} library to train a classifier and perform model inference \cite{plasticc_vrushabh}.

		\subsection{Predictive Analytics in Industrial IoT}
\begin{figure}
					\centering
					\includegraphics[width=\textwidth]{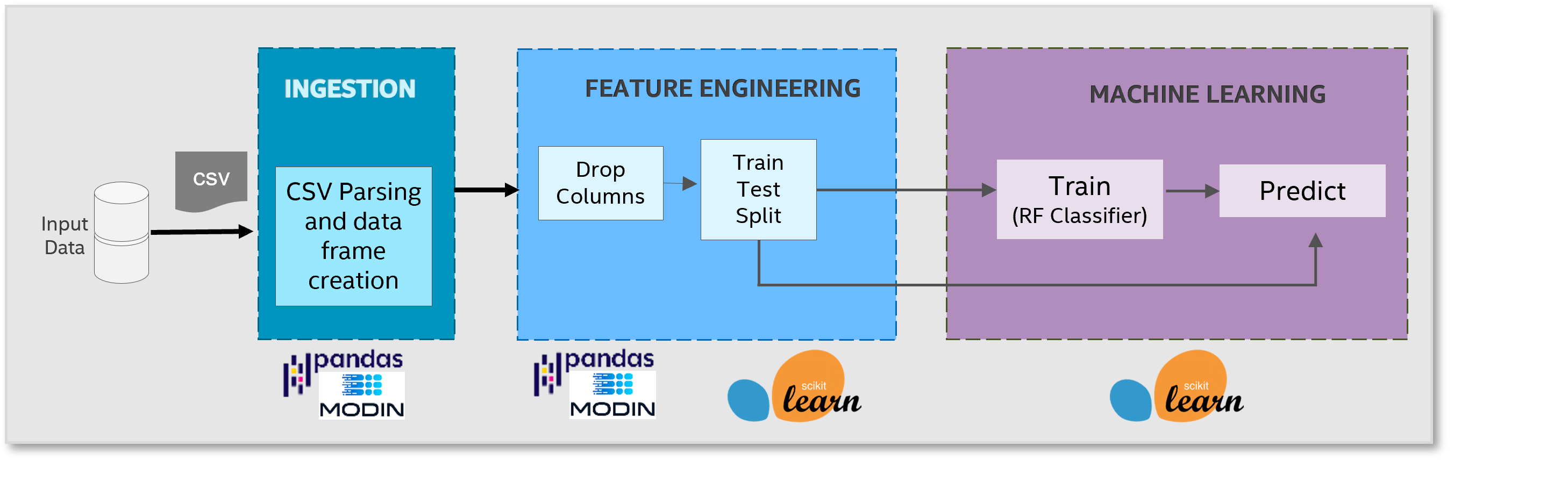}
					\caption{Pipeline for predictive analytics in industrial IoT..}
					\label{fig:fig4}
				\end{figure}

		This is an E2E unsupervised learning use-case in industrial IoT that predicts internal failures during manufacturing \cite{iiot_meena}, thereby helping to maintain the quality and performance of the production line (Figure \ref{fig:fig4}). The workflow consists of reading measurements from a CSV file and cleaning them to include only the necessary features. The highly optimized Intel Distribution of Modin \cite{modindist} is used for this step. The random forest classifier from Intel Extension for Scikit-learn \cite{sklearn-bench, sklearn-gettingstarted} is used to generate the model.

		\subsection{Document Level Sentiment Analysis (DLSA)}
\begin{figure}
					\centering
					\includegraphics[width=\textwidth]{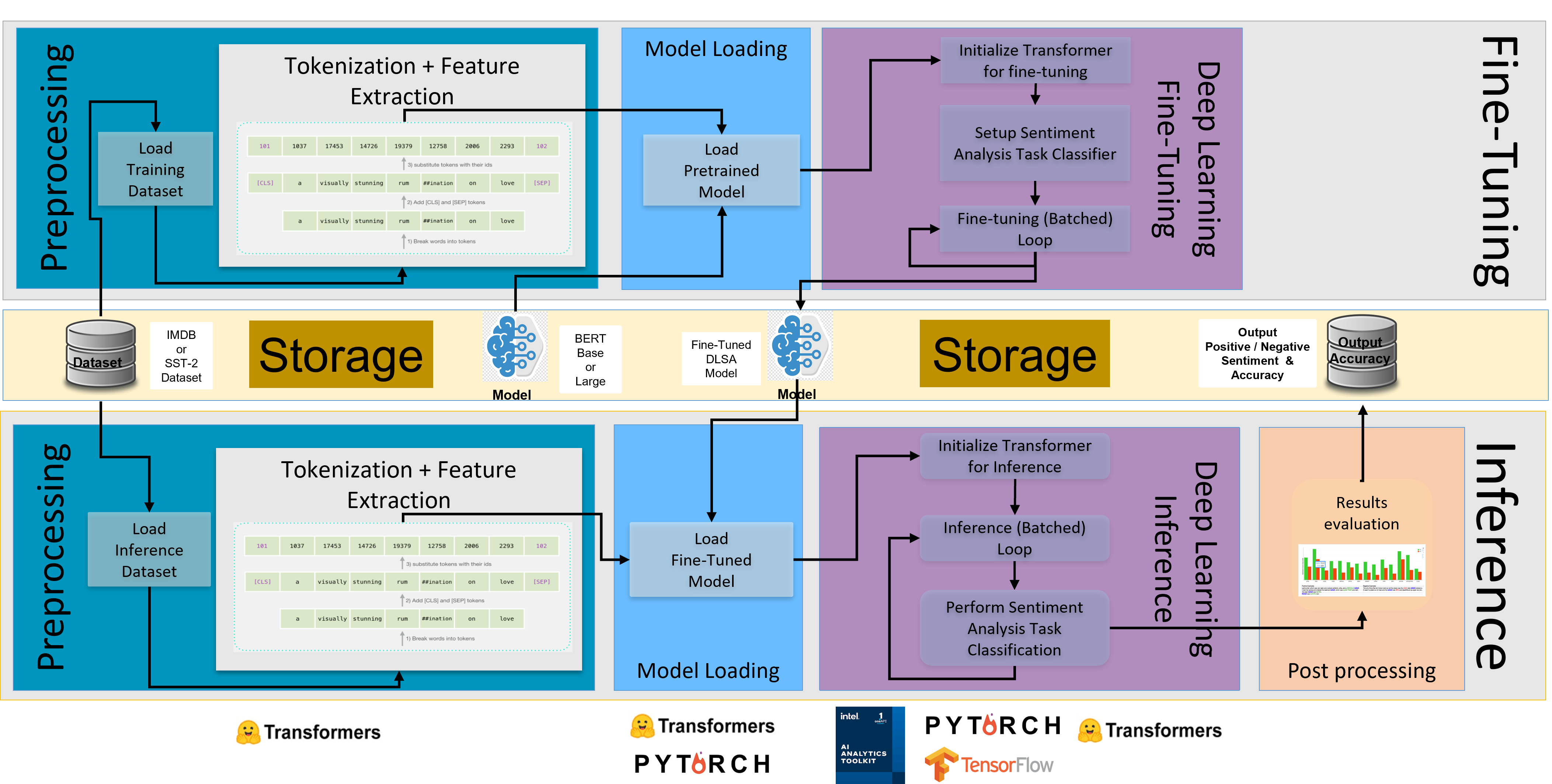}
					\caption{Document level sentiment analysis pipeline.}
					\label{fig:fig5}
				\end{figure}

		Document Level Sentiment Analysis workflow is an E2E deep learning workflow to analyze the sentiment of input documents containing English sentences or paragraphs. This workflow uses Bert-based transformer models \cite{bertpaper} such as Bert-Large or Distilbert \cite{distilbert} available within Huggingface Model Hub \cite{huggingfacebert} for Masked-Language-Modelling task. These models are pretrained on large corpus of English text and then finetuned using movie reviews datasets such as SST-2 \cite{socher-etal-2013-recursive} and IMDB \cite{maas2011} for sentiment analysis task. The DLSA inference workflow, shown in Figure \ref{fig:fig5} is a reference NLP pipeline for document-level sentiment analyses and consist of loading data, tokenizing and feature extraction, loading model and inference.

		\subsection{Deep Interest Evolution Network (DIEN) Recommendation System}
\begin{figure}
					\centering
					\includegraphics[width=\textwidth]{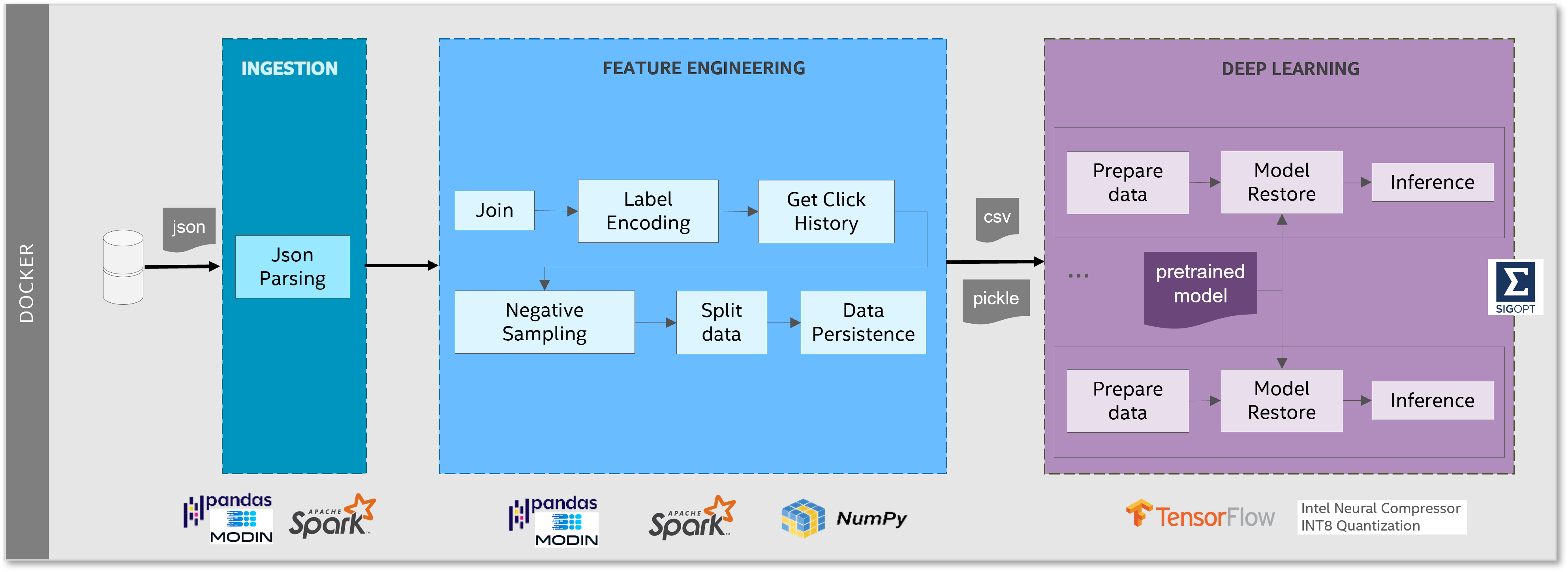}
					\caption{ E2E DIEN recommendation system pipeline.}
					\label{fig:fig6}
				\end{figure}

		The DIEN workflow is being used for Click-through rate (CTR) prediction, with the goal to estimate the probability of a user clicking on the item \cite{dienpaper}. CTR is heavily used in the advertising system and is a big driver to generate revenues for banking, retails and enterprises, and governments. In inference system, usually a batch of 10k to 100M users  data (ranging from few GBs to TBs) is preprocessed and then fed into a deep neural network recommendation model for inference of users behaviors such as clicks and purchases. In our DIEN workflow, during preprocessing, json input is parsed into dataframes, and feature engineering tasks are further optimized to reduce serial code and intermediate data. Finally, inference is done and CTR is predicted by the model (see Figure \ref{fig:fig6}).

		\subsection{Video Streamer}
\begin{figure}
					\centering
					\includegraphics[width=\textwidth]{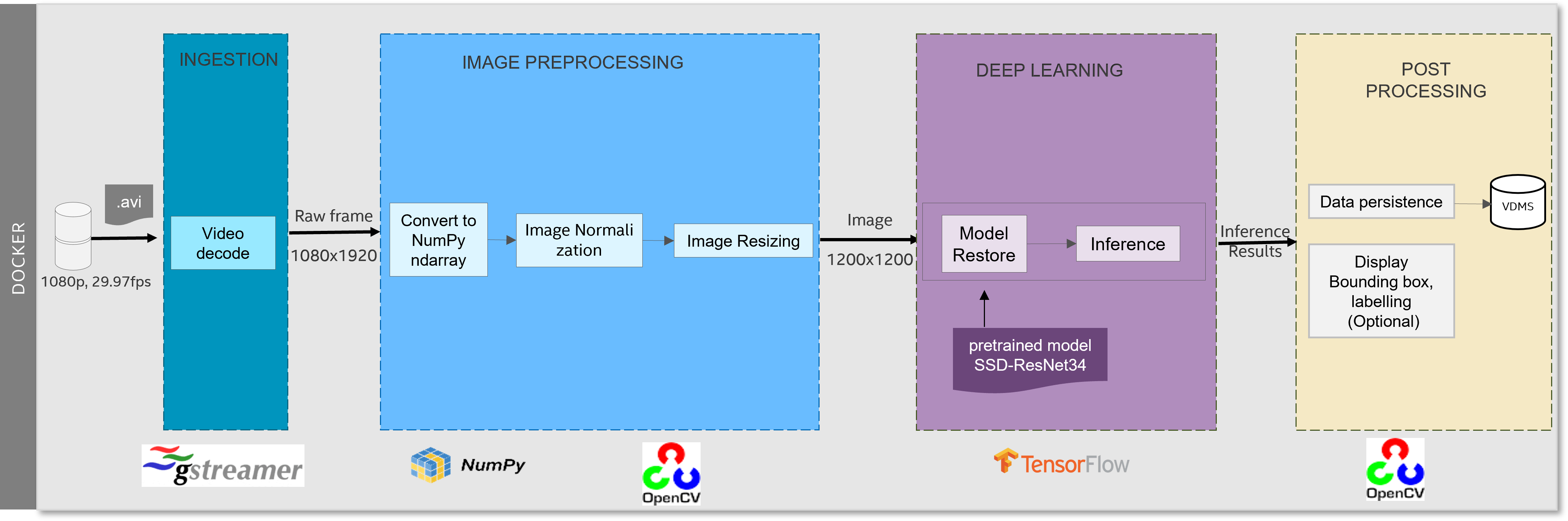}
					\caption{Video streamer application pipeline.}
					\label{fig:fig7}
				\end{figure}

		The video streamer pipeline (Figure \ref{fig:fig7}) is designed to mimic real-time video analytics. Real-time data is provided to an inference endpoint that executes single shot object detection. The metadata created during inference is then uploaded to a database for curation. The pipeline is built upon \verb*|Gstreamer| \cite{gstreamer}, \verb*|TensorFlow|, and \verb*|OpenCV|. The input video is decoded by \verb*|Gstreamer| into images on a frame-by-frame basis. Then, the \verb*|Gstreamer| buffer is converted into a \verb*|NumPy| array. \verb*|TensorFlow| does image normalization and resizing followed by object detection with a pretrained \verb*|SSD-ResNet34| model \cite{ssd_liu2016}. Finally, the results of bounding box coordinates and class labels are uploaded to a database.

		\subsection{Anomaly Detection}
\begin{figure}
						\centering
						\includegraphics[width=\textwidth]{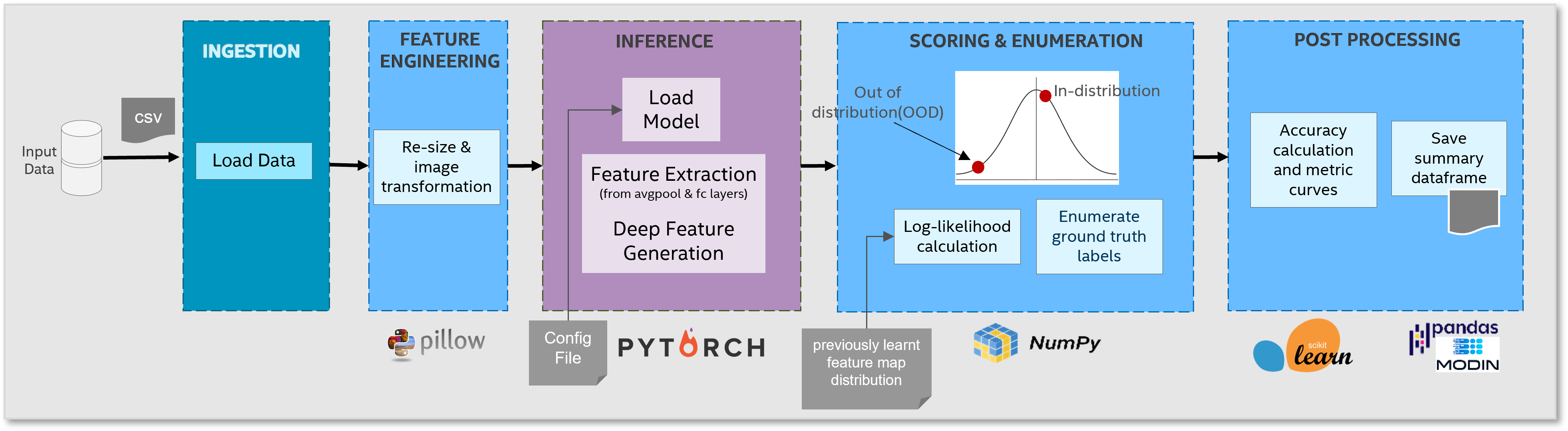}
						\caption{ Anomaly detection pipeline.}
						\label{fig:fig8}
					\end{figure}

		The objective of anomaly detection is to analyze images of parts being manufactured on an industrial production line using deep neural network and probabilistic modeling to identify rare defects (Figure \ref{fig:fig8}). As an out-of-distribution solution, a model of normality is learned over feature maps of the final few layers from normal data in an unsupervised manner. Deviations from the models are flagged as anomalies. Prior to learning distribution, the dimension of the feature space is reduced using PCA to prevent matrix singularities and rank deficiencies from arising while estimating the parameters of the distribution \cite{iiot_meena}.
		
		\subsection{Face Recognition}
\begin{figure}
					\centering
					\includegraphics[width=\textwidth]{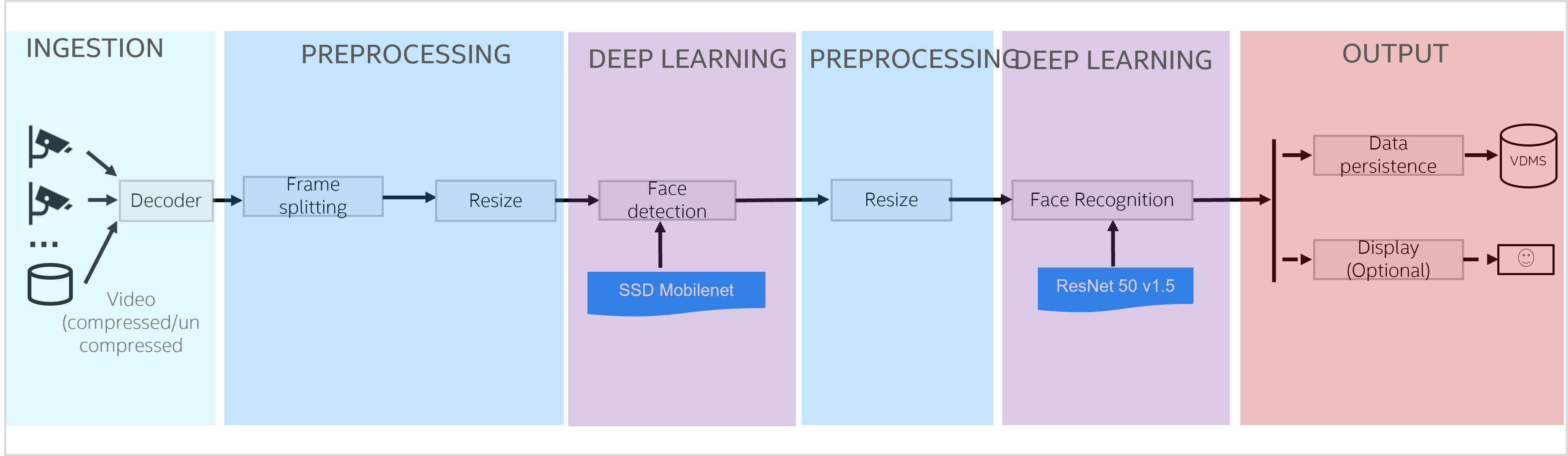}
					\caption{ Face recognition pipeline.}
					\label{fig:fig9}
				\end{figure}

		This E2E pipeline performs real-time face recognition by cascading two out-of-the-box, pretrained models: SSD Mobilenet \cite{mobilenetv2} and Resnet50v1.5 \cite{resnet50_he2015, resnet50v15_intel} (Figure \ref{fig:fig9}). The input from the camera, as compressed or uncompressed video, undergoes frame splitting and resizing. Each frame is then fed to the detection model (SSD Mobilenet), which performs object detection. The NMS bounding boxes are then fed to the recognition model (ResNet50v1.5) to recognize the faces. The output frames with the facial recognitions can either be displayed or saved in databases.
	
\section{Efficient-AI: E2E Optimization Strategies}
\label{sec:efficientai}
\begin{figure}
					\centering
					\includegraphics[width=\textwidth]{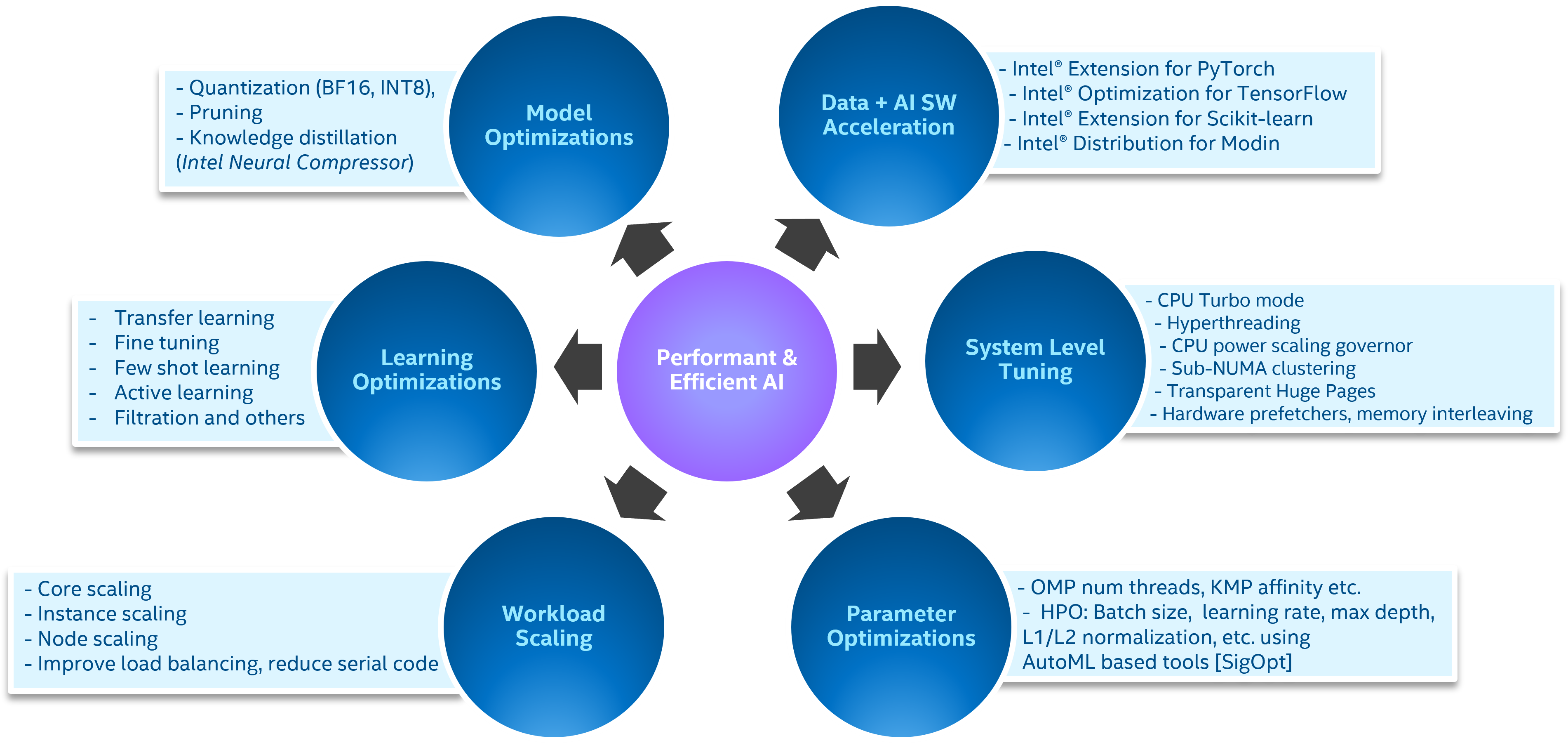}
					\caption{Efficient AI}
					\label{fig:fig10}
				\end{figure}

	E2E performance-efficient AI requires a coherent optimization strategy consisting of AI software acceleration, system-level tuning, hyperparameter and runtime parameter optimizations, and workflow scaling  (Figure \ref{fig:fig10}).  Kang et. al. \cite{kang2020} shown preprocessing might be a bottleneck while executing deep neural models on modern architecture system. By optimizing the preprocessing pipeline they shown a total of 5.9x end-2-end throughput improvement.  All phases (data ingestion, data preprocessing, feature engineering, and model building) need to be holistically addressed to improve user productivity as well as workload performance efficiency \cite{richins2020missing}. 

			\subsection{AI Software Acceleration}
			Intel Distribution of Modin is a multithreaded, parallel and performant data frame library compatible with the pandas API. It performs lightweight, robust data frame and CSV operations and scales efficiently with the number of cores, unlike pandas, providing a significant speedup just by changing a couple of lines of code. Data frame operations across different phases speedup from 1.12x to 30x.
			
			Intel Extension for Scikit-learn accelerates common estimators, transformers, and clustering algorithms in classical ML. Ridge regression training and inference in the Census workload is a DGEMM-based memory-bound algorithm that takes advantage of Intel Extension for Scikit-learn’s vectorization, cache-friendly blocking, and multithreading to efficiently use of multiple CPU cores.
			
			Intel-optimized  and CatBoost libraries provide efficient parallel tree boosting. The XGBoost kernels are optimized for cache efficiency, remote memory latency, and memory access patterns on Intel\textsuperscript{\textregistered} processors.

					\subsubsection{Intel\textsuperscript{\textregistered} Extension for PyTorch (IPEX)}
					IPEX improves PyTorch performance on Intel processors. With IPEX, the Anomaly Detection and DLSA pipelines take advantage of DL Boost. Intel optimizations for TensorFlow \cite{tfperf} is powered by Intel oneAPI Deep Neural Network Library (oneDNN), which includes convolution, normalization, activation, inner product, and other primitives vectorized using Intel AVX-512 instructions. The DIEN, face recognition, and video streamer applications leverage Intel optimizations for TensorFlow to enable scalable performance on Intel processors through vectorization and optimized graph operations (e.g., ops fusion, batch normalization).

			\subsection{Model Optimizations}
			Quantization facilitates conversion of high precision data (32-bit floating point, FP32) to lower precision (8-bit integer, INT8), which enables critical operations such as convolution and matrix multiplication to be performed significantly faster with little to no loss in accuracy. INC automatically optimizes low-precision recipes for DL models and calibrates them to achieve optimal performance and memory usage with expected accuracy criteria. DLSA and video streamer applications achieved up to 4x speedup from INT8 quantization alone (Table 2).
			
						\begin{table}
							\centering
							\includegraphics[width=\textwidth]{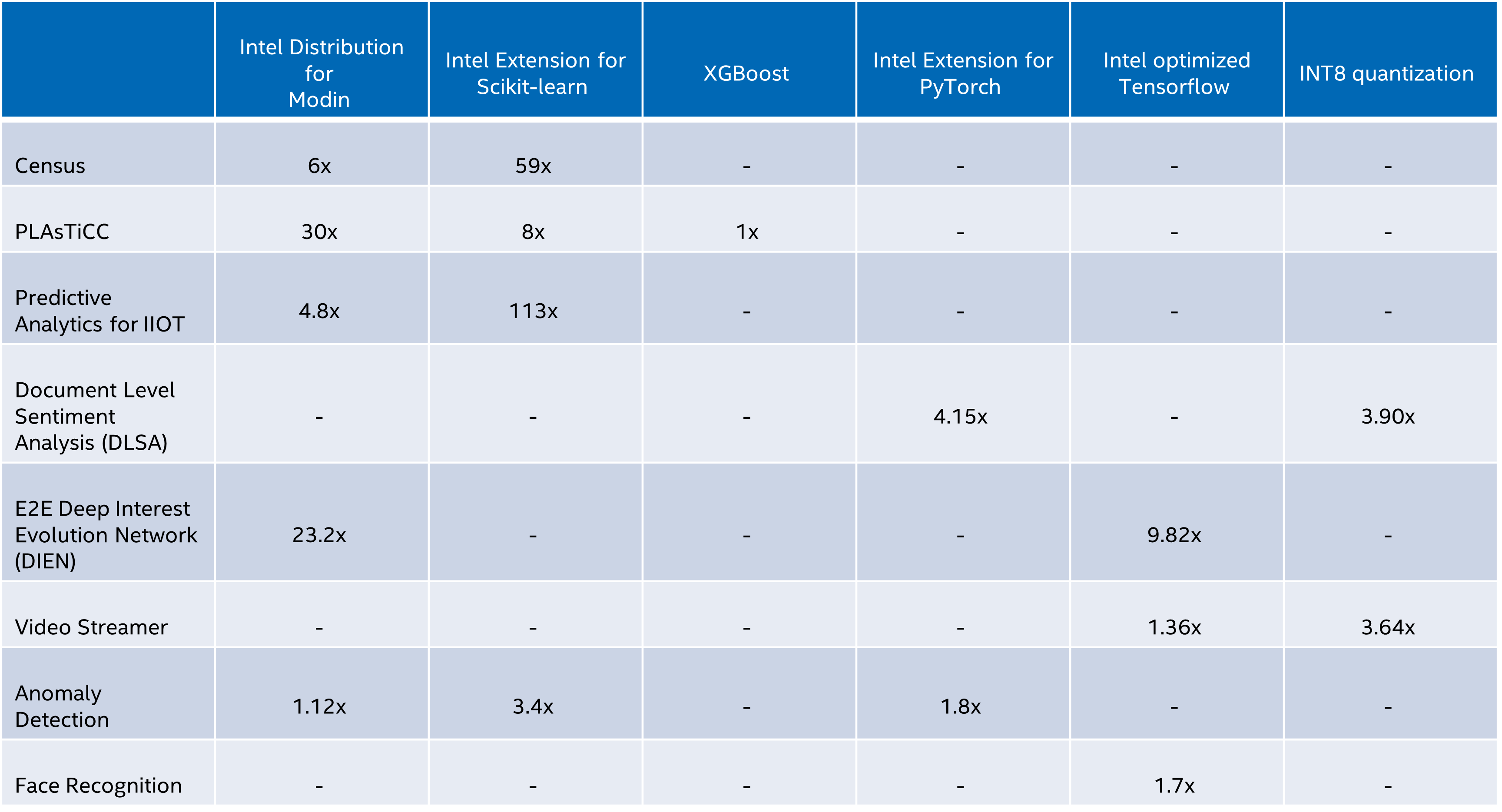}
							\caption{Performance improvement from software optimizations and quantization
								for E2E AI applications.}
							\label{tab:Table2}
						\end{table}

			\subsection{Parameter Optimizations}
			The SigOpt model development platform makes it easy to track runs, visualize training, and scale hyperparameter optimization for any pipeline while tuning for objectives like maximum throughput at threshold accuracy and/or latency levels. With SigOpt’s multi-objective optimization, we can easily obtain insights on the best configurations of the AI pipeline, showing the optimal performance summary and analysis. In the case of PLAsTiCC, ‘accuracy’ and ‘timing’ metrics were optimized while the model hyperparameters (like the number of parallel threads for XGBoost, number of trees, learning rate, max depth, L1/L2 normalization, etc.) were computed in order to achieve the objective \cite{hptuning}. In DLSA, number of inferences instances and batch size are tuned to achieve high E2E throughput.
			
			Run-time options in TensorFlow also make a big performance impact. It is recommended to control the parallelism within an operation like matrix multiplication or reduction so as to schedule the tasks within a threadpool by setting intra\_op\_parallelism\_threads equal to the number of available physical cores and in contrast running operations that are independent in the TensorFlow graph concurrently by setting inter\_op\_parallelism\_threads equal to the number of sockets. Data layout, OpenMP and NUMA controls are also available to tune the performance even further \cite{tfperf}.

			\subsection{Workload Scaling}
			Multi-instance execution allows parallel streams of the application to be executed on a single Intel\textsuperscript{\textregistered} Xeon\textsuperscript{\textregistered} scalable server. The advantage is demonstrated during anomaly detection where several cameras can be deployed to detect defects at different stages of the manufacturing pipeline, 10 such streams processing over the standard 30 FPS on a ResNet50 model can be serviced by a single 3rd Gen Intel\textsuperscript{\textregistered} Xeon\textsuperscript{\textregistered} Scalable processor. Similarly, E2E DIEN runs with one core/instance with 40 inference instances per socket while DLSA and DL pipelines run four cores/instance to eight cores/instance with ten inference streams to five inference streams per socket. This is a unique advantage of CPUs with their large memory capacity.
			
			System level tuning is available in the BIOS to improve efficiency. Tuning knobs controlling hyperthreading, CPU power scaling governors, NUMA optimizations, hardware prefetchers and more can be explored to obtain best performance.

\section{Results and Summary}		
\begin{figure}
								\centering
								\includegraphics[width=\textwidth]{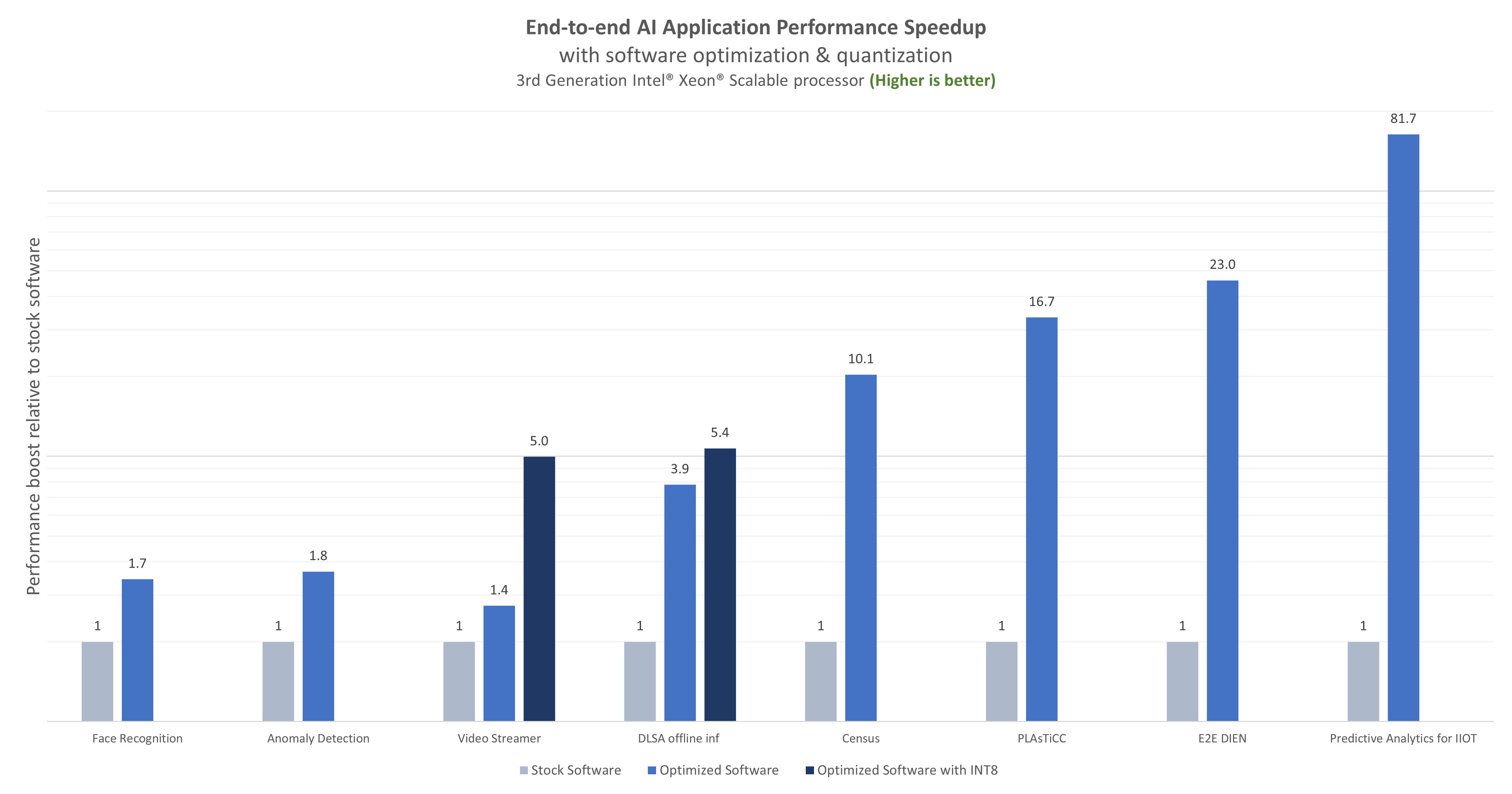}
								\caption{Efficient AI}
								\label{fig:fig11}
							\end{figure}

Preprocessing and postprocessing stages in Artificial Intelligence or Deep learning  pipelines sometimes takes significant amount of time and need to be optimized properly to achieve the overall performance boost. We incorporated several optimization strategies across softwares, systems, hardwares, data ingestion, model building phase, and hyperparameters tuning. With the help of eight pipelines spanning across multiple domains, we were able to achieve E2E performance speedup ranging from 1.8x to 81.7x on Intel\textsuperscript{\textregistered} Xeon\textsuperscript{\textregistered} processors (see Figure ~\ref{fig:fig11}).
			
\paragraph{Configuration}

All the performance was measured on a single-node, dual-socket 3rd Generation Intel\textsuperscript{\textregistered} Xeon\textsuperscript{\textregistered} Scalable 8380 processor (except DIEN and DLSA), 40 cores per socket. DIEN and DLSA were measured on 3rd Generation Intel\textsuperscript{\textregistered} Xeon\textsuperscript{\textregistered} Scalable 6348 processors. The Turbo mode was set \verb+enabled+, and hyperthreading was set to \verb+disabled+. Other configuration are as follows:
BIOS: \verb+SE5C620.86B.01.01.0003.2104260124+, kernel: \verb+5.13.0-28-generic+, OS: \verb+Ubuntu 21.10+, Memory: \verb+16 slots/32GB DIMMs/3200MHz+, Storage: \verb+Intel 480GB SSD OS Drive+. Some important software configuration for all the eight AI pipelines are shown in the Table ~\ref{tab:Table3}

\pagebreak
			
			\begin{table}[htbp]
				\centering
				\caption{Software configurations of the AI workloads.}
				\resizebox{\textwidth}{!}{  
				\begin{tabular}{|ll|}
					\toprule
					\rowcolor[rgb]{ .267,  .447,  .769} \textcolor[rgb]{ 1,  1,  1}{\textbf{E2E pipeline}} & \textcolor[rgb]{ 1,  1,  1}{\textbf{Software versions}} \\
					\midrule
					\rowcolor[rgb]{ .851,  .882,  .949} Anomaly Detection & Python 3.7.11, torch 1.11.0, torchvision 0.11.3, PyTorch 1.10, numpy 1.22.1, pandas 1.3.5, scikit-learn-intelex 2021.4.0; \\
					\midrule
					Face Recognition & Python 3.7.9, TensorFlow 2.8.0, numpy 1.22.2, opencv-python 4.5.3.56, ffmpy 0.3.0;  \\
					\midrule
					\rowcolor[rgb]{ .851,  .882,  .949} Video Streamer & Python 3.8.12, TensorFlow 2.8.0, opencv-python 4.5.2.54, pillow 8.3.1, gstreamer1.0, vdms 0.0.16; \\
					\midrule
					DLSA offline Inf & Python 3.7.11, PyTorch 1.10, HuggingFace Transformer:4.6.1; \\
					\midrule
					\rowcolor[rgb]{ .851,  .882,  .949} E2E DIEN & Python 3.8.10, Modin 0.12.0, TensorFlow 2.8.0, numpy 1.22.2; \\
					\midrule
					Census & Python 3.9.7, Modin 0.12.0, scikit-learn-intelex 2021.4.0; \\
					\midrule
					\rowcolor[rgb]{ .851,  .882,  .949} PLAsTiCC & Python 3.9.7, Modin 0.12.0, scikit-learn-intelex 2021.4.0, XGBoost 1.5.0; \\
					\midrule
					Predictive Analytics for Industrial IoT & Python 3.9.7, Modin 0.12.0, scikit-learn-intelex 2021.4.0. \\
					\bottomrule
				\end{tabular}%
				}
				\label{tab:Table3}%
			\end{table}%

\bibliographystyle{unsrtnat}  
\bibliography{references}  

\begin{thebibliography}{25}
\providecommand{\natexlab}[1]{#1}
\providecommand{\url}[1]{\texttt{#1}}
\expandafter\ifx\csname urlstyle\endcsname\relax
  \providecommand{\doi}[1]{doi: #1}\else
  \providecommand{\doi}{doi: \begingroup \urlstyle{rm}\Url}\fi

\bibitem[cen({\natexlab{a}})]{censuswl}
{Census workload oneAPI sample code, Intel oneAPI AI Analytics toolkit},
  {\natexlab{a}}.
\newblock URL
  \url{https://github.com/oneapi-src/oneAPI-samples/tree/master/AI-and-Analytics/End-to-end-Workloads/Census}.

\bibitem[Ruggles et~al.()Ruggles, Flood, Goeken, Grover, Meyer, Pacas, and
  Sobek]{ipums_data}
Steven Ruggles, Sarah Flood, Ronald Goeken, Josiah Grover, Erin Meyer, Jose
  Pacas, and Matthew Sobek.
\newblock {IPUMS USA: Version 10.0 [dataset]. Minneapolis, MN: IPUMS, 2020. }.
\newblock URL \url{https://www.ipums.org/projects/ipums-usa/d010.v10.0}.

\bibitem[cen({\natexlab{b}})]{census_meena}
{Performance Optimizations for End-to-End AI Pipelines}, {\natexlab{b}}.
\newblock URL
  \url{https://medium.com/intel-analytics-software/performance-optimizations-for-end-to-end-ai-pipelines-231e0966505a}.

\bibitem[{The PLAsTiCC Team} et~al.(2018){The PLAsTiCC Team}, Allam, Bahmanyar,
  Biswas, Dai, Galbany, Hložek, Ishida, Jha, Jones, Kessler, Lochner, Mahabal,
  Malz, Mandel, Martínez-Galarza, McEwen, Muthukrishna, Narayan, Peiris,
  Peters, Ponder, Setzer, Collaboration, Transients, and
  Collaboration]{plasticcpaper}
{The PLAsTiCC Team}, Tarek Allam, Anita Bahmanyar, Rahul Biswas, Mi~Dai, Lluís
  Galbany, Renée Hložek, Emille E.~O. Ishida, Saurabh~W. Jha, David~O. Jones,
  Richard Kessler, Michelle Lochner, Ashish~A. Mahabal, Alex~I. Malz, Kaisey~S.
  Mandel, Juan~Rafael Martínez-Galarza, Jason~D. McEwen, Daniel Muthukrishna,
  Gautham Narayan, Hiranya Peiris, Christina~M. Peters, Kara Ponder,
  Christian~N. Setzer, The LSST Dark Energy~Science Collaboration, The~LSST
  Transients, and Variable Stars~Science Collaboration.
\newblock The photometric lsst astronomical time-series classification
  challenge (plasticc): Data set, 2018.
\newblock URL \url{https://arxiv.org/abs/1810.00001}.

\bibitem[skl({\natexlab{a}})]{sklearn-bench}
{Machine Learning Benchmarks: benchmarks containing implementations of machine
  learning algorithms across data analytics frameworks}, {\natexlab{a}}.
\newblock URL \url{https://github.com/IntelPython/scikit-learn\_bench}.

\bibitem[Chen and Guestrin(2016)]{Chen-2016}
Tianqi Chen and Carlos Guestrin.
\newblock {XGBoost}.
\newblock In \emph{Proceedings of the 22nd {ACM} {SIGKDD} International
  Conference on Knowledge Discovery and Data Mining}. {ACM}, aug 2016.
\newblock \doi{10.1145/2939672.2939785}.
\newblock URL \url{https://doi.org/10.1145%2F2939672.2939785}.

\bibitem[pla()]{plasticc_vrushabh}
{Optimizing Artificial Intelligence Applications}.
\newblock URL
  \url{https://medium.com/intel-analytics-software/optimizing-artificial-intelligence-applications-1bc22b5d707b}.

\bibitem[Arunachalam et~al.(2022)Arunachalam, Sanghavi, Kaira, and
  Ahuja]{iiot_meena}
Meena Arunachalam, Vrushabh Sanghavi, Samudyatha Kaira, and Nilesh~A. Ahuja.
\newblock {End-to-End Industrial IoT: Software Optimization and Acceleration}.
\newblock \emph{IEEE Internet of Things Magazine}, 5\penalty0 (1):\penalty0
  48--53, 2022.
\newblock \doi{10.1109/IOTM.005.2100196}.

\bibitem[mod()]{modindist}
{Intel Distribution of Modin}.
\newblock URL
  \url{https://www.intel.com/content/www/us/en/developer/tools/oneapi/distribution-of-modin.html}.

\bibitem[skl({\natexlab{b}})]{sklearn-gettingstarted}
{Getting started with Intel Extension for Scikit-learn}, {\natexlab{b}}.
\newblock URL
  \url{https://www.intel.com/content/www/us/en/developer/articles/guide/intel-extension-for-scikit-learn-getting-started.html}.

\bibitem[Devlin et~al.(2018)Devlin, Chang, Lee, and Toutanova]{bertpaper}
Jacob Devlin, Ming-Wei Chang, Kenton Lee, and Kristina Toutanova.
\newblock {BERT: Pre-training of Deep Bidirectional Transformers for Language
  Understanding}, 2018.
\newblock URL \url{https://arxiv.org/abs/1810.04805}.

\bibitem[Sanh et~al.(2019)Sanh, Debut, Chaumond, and Wolf]{distilbert}
Victor Sanh, Lysandre Debut, Julien Chaumond, and Thomas Wolf.
\newblock {DistilBERT, a distilled version of BERT: smaller, faster, cheaper
  and lighter}, 2019.
\newblock URL \url{https://arxiv.org/abs/1910.01108}.

\bibitem[hug()]{huggingfacebert}
{HuggingFace Bert Models}.
\newblock URL \url{https://huggingface.co/models?filter=bert}.

\bibitem[Socher et~al.(2013)Socher, Perelygin, Wu, Chuang, Manning, Ng, and
  Potts]{socher-etal-2013-recursive}
Richard Socher, Alex Perelygin, Jean Wu, Jason Chuang, Christopher~D. Manning,
  Andrew Ng, and Christopher Potts.
\newblock Recursive deep models for semantic compositionality over a sentiment
  treebank.
\newblock In \emph{Proceedings of the 2013 Conference on Empirical Methods in
  Natural Language Processing}, pages 1631--1642, Seattle, Washington, USA,
  October 2013. Association for Computational Linguistics.
\newblock URL \url{https://www.aclweb.org/anthology/D13-1170}.

\bibitem[Maas et~al.(2011)Maas, Daly, Pham, Huang, Ng, and Potts]{maas2011}
Andrew~L. Maas, Raymond~E. Daly, Peter~T. Pham, Dan Huang, Andrew~Y. Ng, and
  Christopher Potts.
\newblock {Learning Word Vectors for Sentiment Analysis}.
\newblock In \emph{Proceedings of the 49th Annual Meeting of the Association
  for Computational Linguistics: Human Language Technologies}, pages 142--150,
  Portland, Oregon, USA, June 2011. Association for Computational Linguistics.
\newblock URL \url{http://www.aclweb.org/anthology/P11-1015}.

\bibitem[Zhou et~al.(2018)Zhou, Mou, Fan, Pi, Bian, Zhou, Zhu, and
  Gai]{dienpaper}
Guorui Zhou, Na~Mou, Ying Fan, Qi~Pi, Weijie Bian, Chang Zhou, Xiaoqiang Zhu,
  and Kun Gai.
\newblock {Deep Interest Evolution Network for Click-Through Rate Prediction},
  2018.
\newblock URL \url{https://arxiv.org/abs/1809.03672}.

\bibitem[gst()]{gstreamer}
{gstreamer: Open Source Multimedia Framework}.
\newblock URL \url{https://gstreamer.freedesktop.org/}.

\bibitem[Liu et~al.(2016)Liu, Anguelov, Erhan, Szegedy, Reed, Fu, and
  Berg]{ssd_liu2016}
Wei Liu, Dragomir Anguelov, Dumitru Erhan, Christian Szegedy, Scott Reed,
  Cheng-Yang Fu, and Alexander~C. Berg.
\newblock {SSD}: Single shot {MultiBox} detector.
\newblock In \emph{Computer Vision {\textendash} {ECCV} 2016}, pages 21--37.
  Springer International Publishing, 2016.
\newblock \doi{10.1007/978-3-319-46448-0_2}.
\newblock URL \url{https://doi.org/10.1007%2F978-3-319-46448-0_2}.

\bibitem[Sandler et~al.(2018)Sandler, Howard, Zhu, Zhmoginov, and
  Chen]{mobilenetv2}
Mark Sandler, Andrew Howard, Menglong Zhu, Andrey Zhmoginov, and Liang-Chieh
  Chen.
\newblock {MobileNetV2: Inverted Residuals and Linear Bottlenecks}.
\newblock 2018.
\newblock \doi{10.48550/ARXIV.1801.04381}.
\newblock URL \url{https://arxiv.org/abs/1801.04381}.

\bibitem[He et~al.(2015)He, Zhang, Ren, and Sun]{resnet50_he2015}
Kaiming He, Xiangyu Zhang, Shaoqing Ren, and Jian Sun.
\newblock {Deep Residual Learning for Image Recognition}, 2015.
\newblock URL \url{https://arxiv.org/abs/1512.03385}.

\bibitem[res()]{resnet50v15_intel}
{Intel ResNet 50v1.5 models}.
\newblock URL
  \url{https://github.com/IntelAI/models/tree/master/benchmarks/image_recognition/tensorflow/resnet50v1_5}.

\bibitem[Kang et~al.(2020)Kang, Mathur, Veeramacheneni, Bailis, and
  Zaharia]{kang2020}
Daniel Kang, Ankit Mathur, Teja Veeramacheneni, Peter Bailis, and Matei
  Zaharia.
\newblock {Jointly Optimizing Preprocessing and Inference for DNN-based Visual
  Analytics}, 2020.
\newblock URL \url{https://arxiv.org/abs/2007.13005}.

\bibitem[Richins et~al.(2020)Richins, Doshi, Blackmore, Nair, Pathapati, Patel,
  Daguman, Dobrijalowski, Illikkal, Long, et~al.]{richins2020missing}
Daniel Richins, Dharmisha Doshi, Matthew Blackmore, Aswathy~Thulaseedharan
  Nair, Neha Pathapati, Ankit Patel, Brainard Daguman, Daniel Dobrijalowski,
  Ramesh Illikkal, Kevin Long, et~al.
\newblock {Missing the forest for the trees: End-to-end ai application
  performance in edge data centers}.
\newblock In \emph{2020 IEEE International Symposium on High Performance
  Computer Architecture (HPCA)}, pages 515--528. IEEE, 2020.

\bibitem[tfp()]{tfperf}
{Maximize TensorFlow Performance on CPU}.
\newblock URL
  \url{https://www.intel.com/content/www/us/en/developer/articles/technical/maximize-tensorflow-performance-on-cpu-considerations-and-recommendations-for-inference.html}.

\bibitem[hpt()]{hptuning}
{Optimizing Artificial Intelligence Applications with Hyperparameter Tuning and
  Optimized Software}.
\newblock URL
  \url{https://medium.com/intel-analytics-software/optimizing-artificial-intelligence-applications-1bc22b5d707b}.

\end{thebibliography}

\end{document}